\documentclass{article}

\usepackage{microtype}
\usepackage{graphicx}
\usepackage{booktabs} 
\usepackage{multirow}
\usepackage{makecell}
\usepackage{arydshln}   
\usepackage{graphicx}
\usepackage{graphics}
\usepackage{bbm}
\usepackage{balance}
\usepackage{algorithm}
\usepackage{color}
\usepackage{amsmath}
\usepackage{amssymb}
\usepackage{textcomp}
\usepackage{color}
\usepackage[dvipsnames]{xcolor}
\usepackage{color}
\usepackage{subfig}

\definecolor{zred}{RGB}{196, 38, 11}
\definecolor{zblue}{RGB}{41, 52, 190}
\definecolor{zgreen}{RGB}{18, 141, 21}

\definecolor{zptu}{RGB}{18, 141, 21}

\usepackage{hyperref}

\usepackage[accepted]{icml2021}

\icmltitlerunning{Order-Agnostic Cross Entropy for Non-Autoregressive Machine Translation}

\begin{document}

\twocolumn[
\icmltitle{Order-Agnostic Cross Entropy for Non-Autoregressive Machine Translation}



\icmlsetsymbol{equal}{*}

\begin{icmlauthorlist}
\icmlauthor{Cunxiao Du}{smu}
\icmlauthor{Zhaopeng Tu}{tencent}
\icmlauthor{Jing Jiang}{smu}
\end{icmlauthorlist}

\icmlaffiliation{smu}{School of Computing and Information System, Singapore Management University, Singapore. Work was done when Cunxiao Du was under the Rhino-Bird Elite Training Program of Tencent AI Lab.}
\icmlaffiliation{tencent}{Tencent AI Lab, China}

\icmlcorrespondingauthor{Zhaopeng Tu}{zptu@tencent.com}

\icmlkeywords{Machine Learning, ICML}

\vskip 0.3in
]



\printAffiliationsAndNotice{}

\begin{abstract}
We propose a new training objective named {\em order-agnostic cross entropy} (\textsc{OaXE}) for fully non-autoregressive translation (NAT) models. \textsc{OaXE} improves the standard cross-entropy loss to ameliorate the effect of word reordering, which is a common source of the critical multimodality problem in NAT.
Concretely, \textsc{OaXE} removes the penalty for word order errors, and computes the cross entropy loss based on the best possible alignment between model predictions and target tokens.
Since the log loss is very sensitive to invalid references, we leverage cross entropy initialization and loss truncation to ensure the model focuses on a good part of the search space.
Extensive experiments on major WMT benchmarks show that \textsc{OaXE} substantially improves translation performance, setting new state of the art for fully NAT models. Further analyses show that \textsc{OaXE} alleviates the multimodality problem by reducing token repetitions and increasing prediction confidence.
{Our code, data, and trained models are available at \protect\url{https://github.com/tencent-ailab/ICML21_OAXE}.}
\end{abstract}

\section{Introduction}
\label{intro}

Non-autoregressive translation~\citep[NAT,][]{NAT,libovicky-helcl-2018-end} has received increasing attention for its efficient decoding by predicting every target token in parallel. 
However, such advantage comes at the cost sacrificing translation quality due to the {\em multimodality} problem.
There exist many possible translations of the same sentence, while vanilla NAT models may consider them at the same time due to the independent predictions,
which leads to multi-modal outputs in the form of token repetitions~\cite{NAT}.

A number of recent efforts have explored ways to improve the NAT models' ability to handle multimodality.
One thread of research relaxes the fully non-autoregressive restriction and iteratively refine the generated outputs with K decoding passes~\cite{iterativerefine,gu2019levenshtein}. However, the iterative refinement method improves translation quality by sacrificing the primary benefit of NAT models -- fast inference~\cite{kasai2021deep}.

\begin{figure*}[t]
    \centering
    \subfloat[Standard XE]{
    \includegraphics[height=0.11\textwidth]{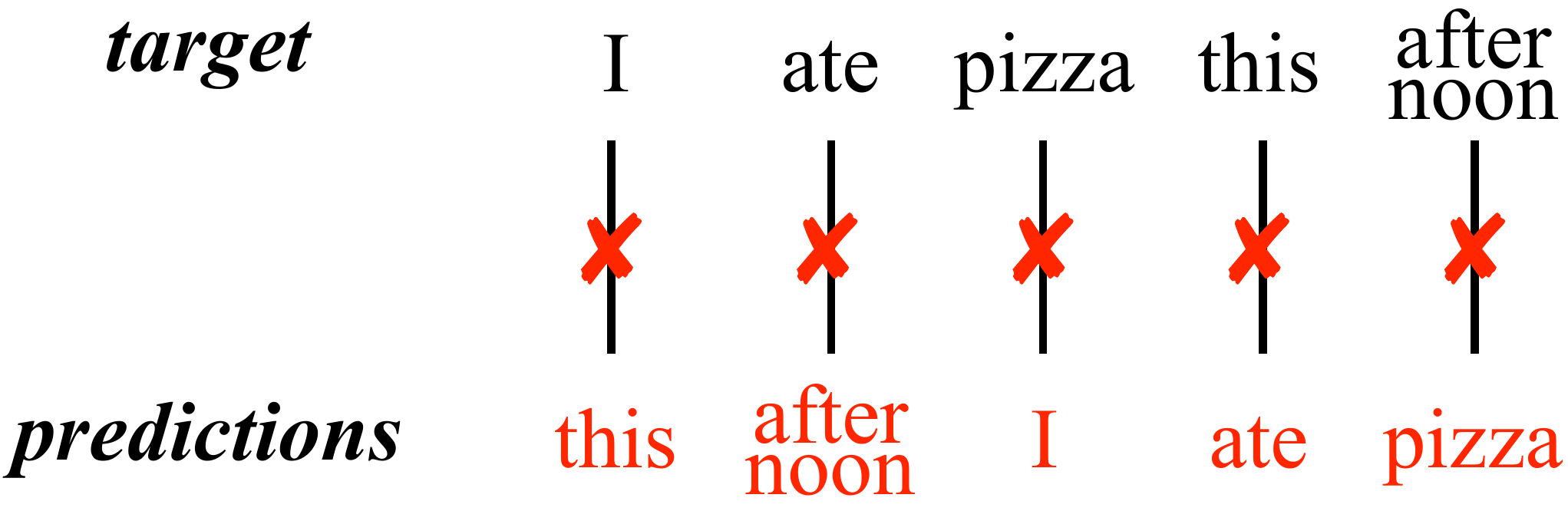} \label{fig:loss-ce}
    } \hfill
    \subfloat[Aligned XE]{
    \includegraphics[height=0.11\textwidth]{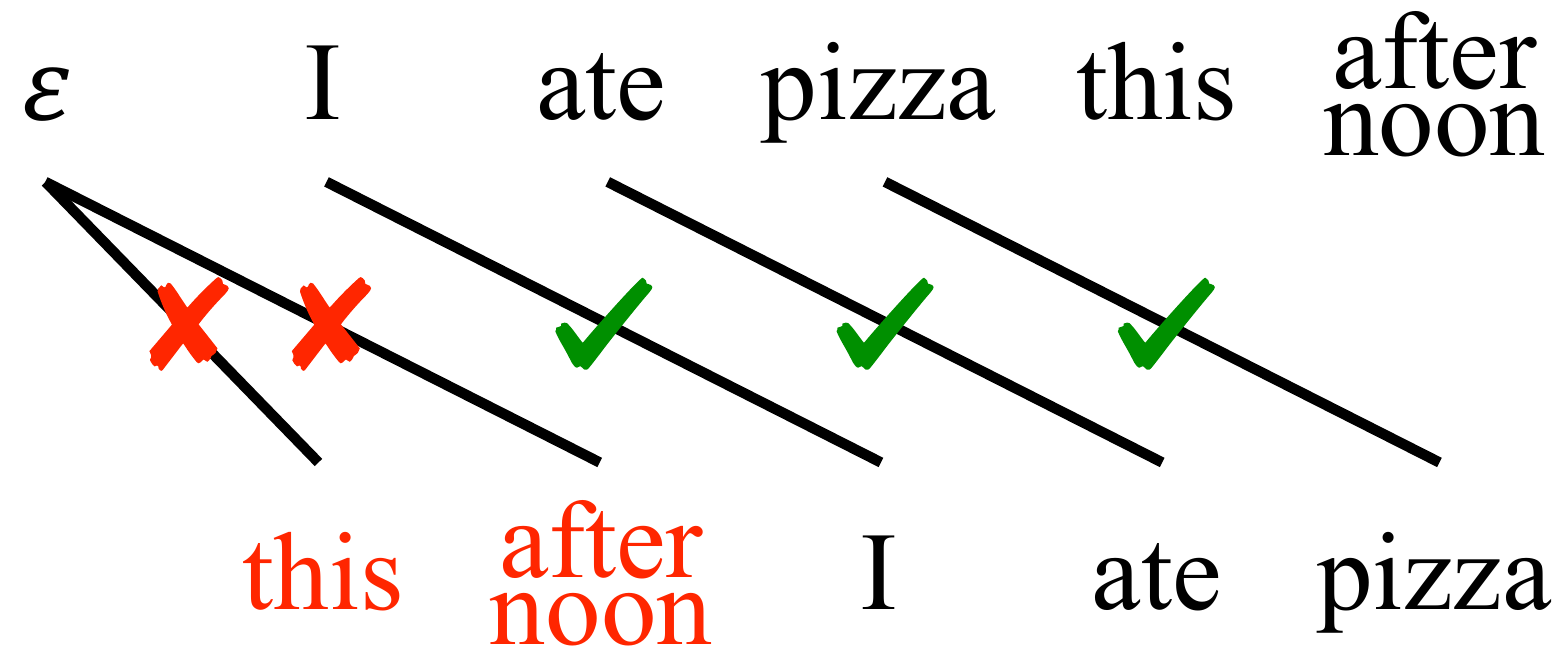} \label{fig:loss-axe}
    } \hfill
    \subfloat[{\em Order-Agnostic} XE]{
    \includegraphics[height=0.11\textwidth]{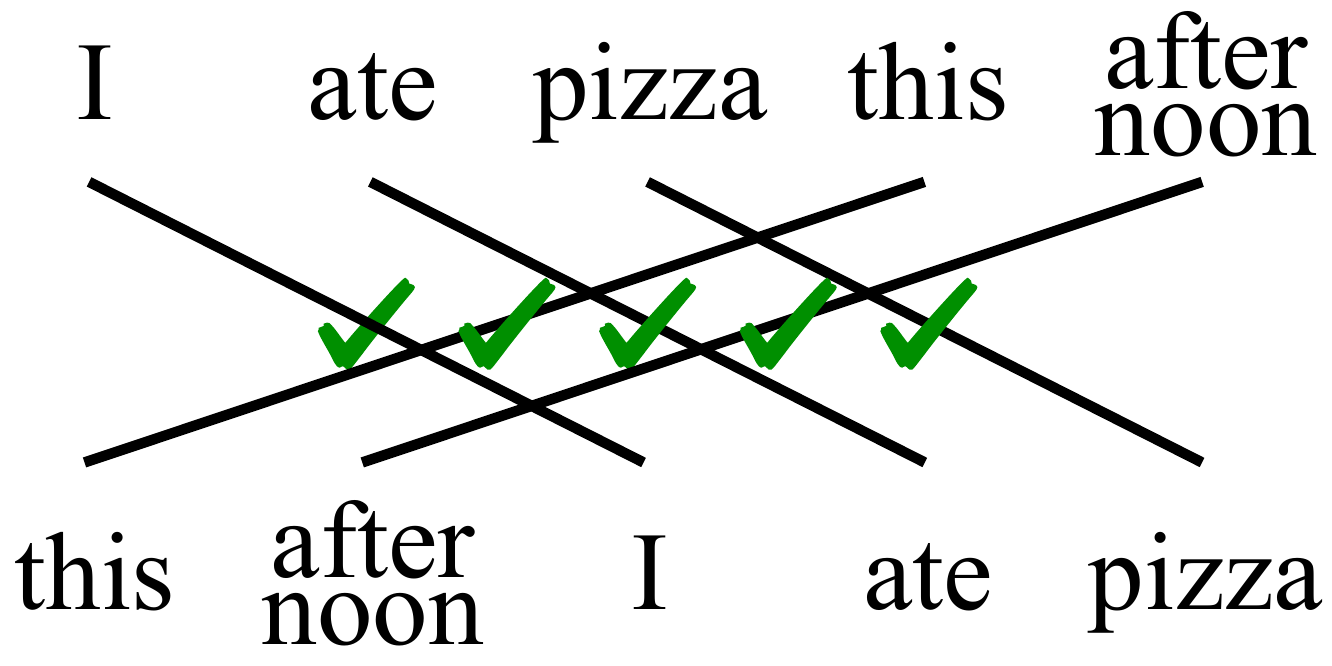} \label{fig:loss-oace}
    }
    \caption{Illustration of the loss functions: (a) the standard XE~\cite{NAT,maskp} where a penalty is incurred for every out-of-position prediction, (b) a relaxed AXE~\cite{axe} that assigns loss based on {\em the best possible monotonic alignment} between model predictions and target tokens, and (c) our proposed Order-Agnostic XE that assigns loss based on {\bf the best possible alignment} between predictions and target.  
    We highlight in red color the predictions that are penalized by the loss function.}
    \label{fig:loss}
\end{figure*}

Another thread of work focuses on improving {\em fully} NAT models (i.e., one decoding pass) to maintain their advantage of decoding efficiency. Along this direction, recent works have incorporated approaches to improving the standard cross-entropy loss to ameliorate the effect of multimodality.
~\citet{axe} state that modeling word order is difficult for NAT, since the model cannot condition on its previous predictions like its autoregressive counterpart.
Starting from this intuition, they propose a new training loss that {\bf softens} the penalty for word order errors, which assigns loss based on the best possible {monotonic} alignment between target tokens and model predictions. Similarly,~\citet{libovicky-helcl-2018-end, imputer} marginalize all the {monotonic} alignments between target and predictions to compute loss. 
While both approaches improve NAT performance by relaxing the word order restriction based on the {\bf monotonic} alignment assumption, they cannot handle {\bf word reordering} -- a common source of multimodality problem (see Figure~\ref{fig:loss}).
In response to this problem, we propose to {\bf remove} the penalty for word order errors, which makes the loss more compatible with parallel decoding in NAT.

To achieve this more relaxed loss, we introduce order-agnostic cross entropy (\textsc{OaXE}), a new training objective function that computes the cross entropy loss based on the best possible alignment between model predictions and target tokens. Compared with previous studies~\cite{axe,imputer}, \textsc{OaXE} guides NAT models to focus on lexical matching, which is at the core of machine translation tasks.
We use {Hungarian algorithm} to efficiently implement \textsc{OaXE} (e.g., 7 lines of core code, see Appendix~\ref{sec:app-pseudocode}), which makes our new loss accommodate existing NAT models seamlessly.
The log loss is sensitive to invalid and noisy references~\cite{losstruncation}, and there exists a large number of invalid alignments in the search space, which will hurt the performance of \textsc{OaXE} in the form of log loss. We tackle these challenges by removing invalid alignments via cross entropy initialization to ensure the model focuses on a good part of the search space. We also use loss truncation~\cite{losstruncation} to drop invalid model predictions, which biases the model towards learning only the confident (and likely accurate) predictions.

We first design a synthetic experiment to demonstrate that \textsc{OaXE} better captures the multimodality of word orders.
Extensive experiments on several machine translation benchmarks with different settings show that \textsc{OaXE} substantially improves translation performance over NAT models trained with both cross entropy~\cite{maskp} and aligned cross entropy~\citep[AXE,][]{axe}. \textsc{OaXE} increases training time by 1.36$\times$ (on par with cross entropy), and does not affect inference speed.
Our comprehensive analyses show that \textsc{OaXE} indeed alleviates the critical multimodality problem by reducing token repetitions and increasing prediction confidence. 
\textsc{OaXE} can also improve output fluency by benefiting from its ability to find an appropriate ordering of generated tokens.

\paragraph{Contributions} Our main contributions are as follows:
\begin{itemize}
    \item We demonstrate the necessity of explicitly handling word reordering -- a common source of the critical multimodality problem in NAT.
    \item We propose a new training objective for NAT that removes the penalty for word order errors, which consistently outperforms the standard cross entropy loss on several benchmarking datasets with and without knowledge distillation. 
    \item We achieve new state-of-the-art performances for fully NAT on six machine translation benchmarks: WMT14 En$\Leftrightarrow$De (26.1 and 30.2), WMT16 En$\Leftrightarrow$Ro (32.4 and 33.3), and WMT17 En$\Leftrightarrow$Zh (32.9 and 22.1), which closely match the performance of AT model.
\end{itemize}

\section{Methodology}

In this section, we first briefly introduce the NAT model and its limitation on handling word order (Section~\ref{sec:preliminary}). We then describe the proposed Order-Agnostic Cross Entropy (\textsc{OaXE}) loss, which alleviates the limitation by assigning loss based on the best possible alignment between target tokens and model predictions (Section~\ref{sec:oace}). 
To ease the training of \textsc{OaXE}, we propose two training strategies to remove invalid alignments in the search space (Section~\ref{sec:training}).

\subsection{Preliminaries: Non-Autoregressive Translation}
\label{sec:preliminary}

\paragraph{Cross Entropy (XE)}
Standard NAT models~\cite{NAT} are trained with the vanilla cross entropy loss:
\begin{equation}
    \mathcal{L}_{XE} = -\log P(Y | X) = - \sum_{y_n} \log P(y_n | X), \label{eq:ce}
\end{equation}
where $(X, Y)$ with $Y=\{y_1, \dots, y_N\}$ is a bilingual training example, and $P(y_n | X)$ is calculated independently by the NAT model with parameters $\theta$.
As shown in Figure~\ref{fig:loss-ce}, the standard XE requires a strict match of word order between target tokens and model predictions, thus will heavily penalize all predictions for ``this afternoon I ate pizza'', although it is semantically equivalent to the target.

\paragraph{Aligned Cross Entropy (AXE)}
Several recent efforts improve the standard cross-entropy loss to ameliorate the effect of multimodality of word order~\cite{libovicky-helcl-2018-end,axe,imputer}. Closely related to this work,~\citet{axe} softens the penalty for word order errors based on a {\bf monotonic alignment} assumption. They define an alignment $\alpha$ to be a function that maps target positions to prediction positions, and the conditional AXE is calculated as:
\begin{equation}
    \mathcal{L}_{AXE} = - \sum_{y_n} \log P_\alpha(y_n | X) - \sum_{k \notin \alpha} P_k (\epsilon), \label{eq:axe}
\end{equation}
where the first term is an aligned cross entropy between target and predictions, and the second term is a penalty for unaligned predictions.
As shown in Figure~\ref{fig:loss-axe}, the monotonic assumption of \textsc{AXE} can recover part of the mistakenly penalized predictions (e.g., ``I ate pizza''), but fails to handle reordered predictions (e.g., ``this afternoon'').

\subsection{Order-Agnostic Cross Entropy \textsc{OaXE}}
\label{sec:oace}

\paragraph{Intuition} 

Word reordering -- one of the structural sentence characteristics, is a common source of multimodality.
Two semantically equivalent sentences can vary greatly in how they arrange sentence components. For example, sentences can be in the active or passive form; a temporal adverbial can be placed either at the beginning of a sentence (e.g., ``{\em this afternoon} I ate pizza'') or at the end (e.g., ``I ate pizza {\em this afternoon}'').
~\citet{em} have shown that NAT is hard to fit all the semantically equivalent translations via cross entropy; however, NAT is powerful enough to fit one specific translation. The observation opens up the possibility that an appropriate loss can guide the model to find a possible translation that is semantically equivalent to the target while in a different word order.
Figure~\ref{fig:loss-oace} illustrates the intuition of the proposed \textsc{OaXE} to handle word reordering for NAT models.
\textsc{OaXE} removes the word order restriction of XE, and assigns loss based on the best alignment between target tokens and model predictions.

\paragraph{Formulation and Implementation}
Given one training example $(X, Y)$, each alignment between model predictions and target tokens is an ordering of the set of target tokens $(y_1, \dots, y_N)$ (e.g., $Y^i = \{y_N, y_1, \dots, y_{N-1}\}$). 
Without loss of generality, from here on, {\em we refer to each alignment as an ordering as well}.
Specifically, we define the ordering space ${\bf O}=\{O^1, \dots, O^I\}$ for $Y$, and the \textsc{OaXE} objective is defined as finding the best ordering $O^i$ to minimize the cross entropy loss:
\begin{equation}
    \mathcal{L}_{\textsc{OaXE}} = \underset{O^i \in {\bf O}}{\operatorname{argmin}} \left(-\log P(O^{i}|X) \right), \label{eq:oace}
\end{equation}
where $-\log P(O^{i}|X)$ is the cross entropy loss for ordering $O^i$, which is calculated by Equation~\ref{eq:ce}.
Briefly, after we got the best ordering, we used it to reorder the original reference, which will be used to calculate the loss $-\log P(O^{i}|X)$ using Equation~\ref{eq:ce}.
For example, the original reference ``I ate pizza this afternoon'' will be reordered to ``this afternoon I ate pizza'', which is used to calculate the \textsc{XE} loss.

\begin{figure}[t]
    \centering
    \includegraphics[width=0.4\textwidth]{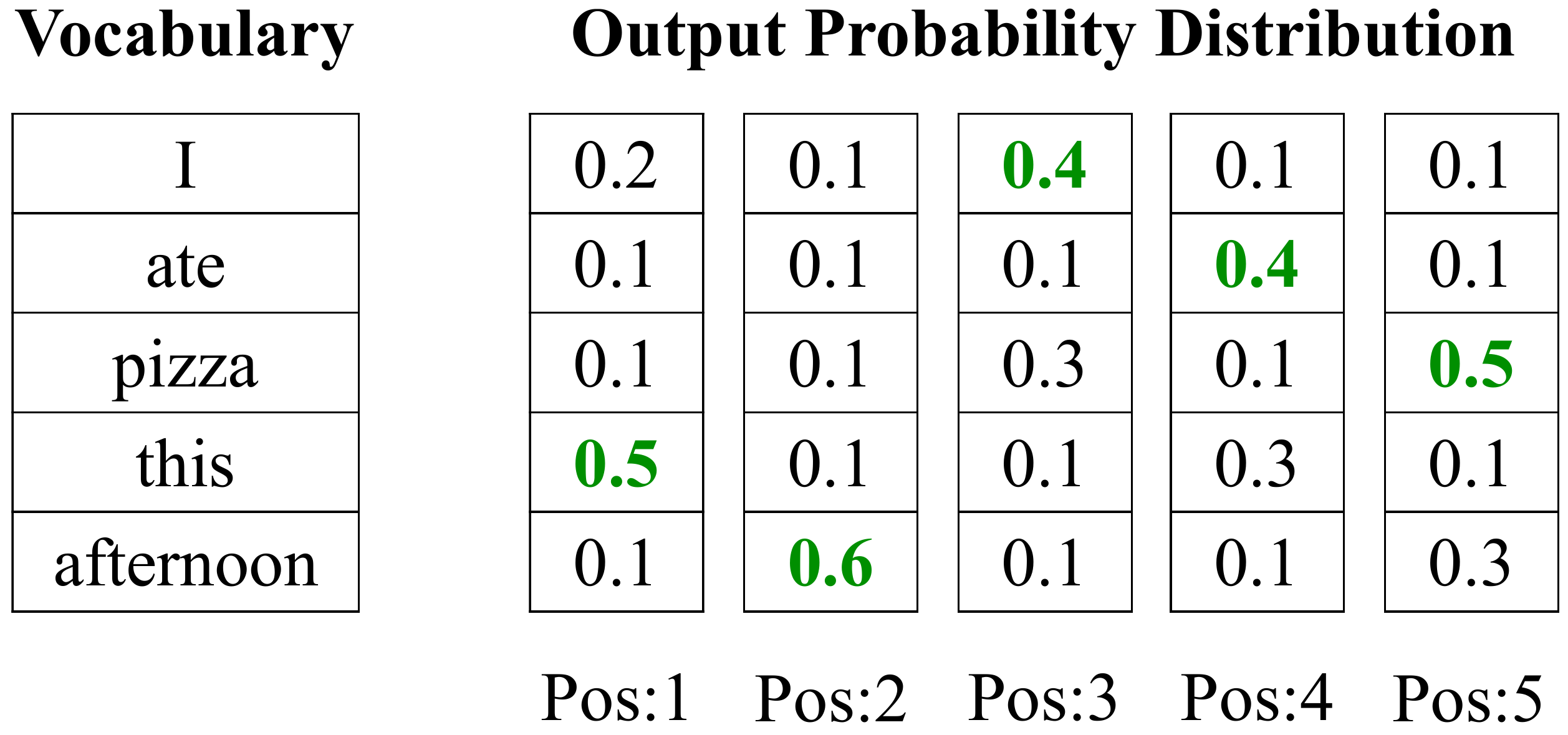}
    \caption{Illustration of bipartite matching to implement the proposed order-agnostic XE loss.
    We only show the probabilities of the target words for better illustration, and highlight in bold the prediction selected by \textsc{OaXE} for each position. The best ordering is ``this afternoon I ate pizza''.}
    \label{fig:bipartite-matching}
\end{figure}

For the target $Y$ with $N$ tokens, there exists $N!$ orderings in the search space $\bf O$, which makes it computationally infeasible to calculate Equation~\ref{eq:oace} explicitly.
In response to this problem, we cast this problem as Maximum Bipartite Matching and leverage the efficient Hungarian algorithm~\cite{kuhn1955hungarian}, which reduces the time complexity from $\mathcal{O}(N!)$ to $\mathcal{O}(N^3)$.
Specifically, as shown in Figure~\ref{fig:bipartite-matching}, we construct the bipartite graph $G=(U, V, E)$ where the first part of vertices $U$ is the set of $N$ positions and the second part of vertices $V$ is the set of $N$ target tokens. Each edge in $E$ is the prediction log probability for the token $y$ in position $n$. 
With the efficient implementation,\footnote{We implemented Hungarian Match with a CPU-version python package \textit{scipy}, which iteratively handles each instance in a mini-batch. This can be accelerated by a GPU-friendly version~\citep[e.g.][]{lopes2019fast}, which we leave for the future work.} 
the training time of the \textsc{OaXE} loss is similar to (about 1.36 times slower than) training with cross entropy loss.\footnote{The training process consists of both the forward pass (e.g. layer and softmax operations) and backward pass (e.g. {\bf loss calculation} and parameter update). Accordingly, although the additional computation of the loss \textsc{OaXE} is $\mathcal{O}(N^3)$, the training speed is only 1.36 times slower.}
Note that \textsc{OaXE} requires no changes to parallel decoding in inference.

\subsection{Training}
\label{sec:training}

A large portion of the $N!$ possible orderings in $\bf O$ are invalid (e.g., ``pizza I this ate afternoon''), and log loss requires that models assign probabilities to all potential reference sequences (i.e., orderings in this study).
\citet{losstruncation} have demonstrated that log loss is not robust to noisy and invalid references, 
since log loss is sensitive to outliers: invalid or noisy references with small probability mass can cause large changes in model behavior.
In response to this problem, we design two strategies to remove invalid references at the levels of both sentence (i.e., invalid orderings) and token (i.e., invalid predictions).

\paragraph{Avoiding Invalid Orderings via XE Initialization}
~\citet{permutations} have shown that the marginal distribution can be used as first order approximation to the joint probability of distributions on permutations. 
As pointed by~\citet{Kasai2020DisCo} and~\citet{maskstrategy}, the NAT model trained with vanilla cross entropy could be view as estimating the marginal distribution $P(y_{n}|X)$ for the $n^{th}$ target token.
Starting from this intuition, we initialize with XE loss to make sure the model can effectively deal with the large search space of orderings.
Such strategy has been successfully applied in sequence-level training for autoregressive models~\cite{Ranzato:2016:ICLR,Shen:2016:MRT,Kong:2019:AAAI} to deal with the large search space of the sequence-level loss. 
Specifically, we design two methods to initialize the \textsc{OaXE}-based NAT with the XE loss.

For the first idea, we start with an initialization pre-trained with the XE loss and slowly deviate from it, instead of starting from a poor random initialization. This ensures that we start off with a much better initialization than random, since now the model focuses on a good part of the search space, which can avoid a large portion of invalid orderings.

The second idea is to introduce \textsc{OaXE} loss during training with an annealing schedule in order to gradually teach the model to remove the order constraint for loss calculation. Specifically, we follow~\citet{bowastarget} to combine XE and \textsc{OaXE} losses:
\begin{equation*}
    \mathcal{L}_{\rm \textsc{joint}} =T_{m} * \mathcal{L}_{\rm \textsc{XE}} + (1-T_{m}) * \mathcal{L}_{\rm \textsc{OaXE}},
\end{equation*}
where $T_{m}$ is the temperature at the $m^{th}$ epoch of training, which progressively decreases to 0:
\begin{equation*}
    T_{m}= \max(0, 1-c^{m - \lambda M}),
\end{equation*}
where $M$ is the total epochs of training, and ($c$, $\lambda$) are pre-defined constant numbers. Accordingly, in the last $(1-\lambda)M$ epochs, only $\textsc{OaXE}$ is used to train the models.

\paragraph{Dropping Invalid Predictions via Loss Truncation}
Even with XE initialization, the model may still find orderings with invalid predictions (e.g., ``I {\em pizza ate} this afternoon''), which are back propagated to update the model parameters. To tackle this problem, we drop the invalid predictions in the searched ordering $\widehat{O}=\{\hat{y}_1, \dots, \hat{y}_N\}$ with loss truncation~\cite{losstruncation}:
\begin{equation*}
    \mathcal{L}_{\textsc{OaXE-T}} = \sum_{\hat{y}_n: P(\hat{y}_n)> \pi} \left( -\log P(y_n | X) \right), \label{eq:oace-t}
\end{equation*}
where $\pi$ is the truncation margin, which is tuned on the validation set.
We only back-propagate the tokens whose probabilities are higher than the margin, which biases the model towards learning only the confident (and likely factually accurate) predictions.
We have empirically found that truncating after XE hotstarting primarily drops invalid predictions, which is consistent with the findings reported by~\citet{losstruncation}.

\section{Synthetic Ordering Experiment}

We first evaluate the effectiveness of the proposed approach to alleviate the multimodality problem in terms of word reordering, which is closely related to the proposed order-agnostic XE loss. For this, we design a synthetic dataset to simulate the word reordering of machine translation tasks, and simplify the other characteristics.

\begin{table}
\centering
\begin{tabular}{ccc}
\toprule
\bf Mode & \bf Source & \bf Target \\
\midrule
Direct    & 1 3 10 7 9 2  &  1 3 10 7 9 2\\
Reverse   & 1 3 10 7 9 2  &  {\em 2 9 7 10 3 1}\\
Flip      & 1 3 10 7 9 2  &  7 9 2 1 3 10\\
Flip-Right-Rev & 1 3 10 7 9 2  &  7 9 2 {\em 10 3 1}\\
Filp-Left-Rev & 1 3 10 7 9 2  &  {\em 2 9 7} 1 3 10\\
\bottomrule
\end{tabular}
\caption{Examples of the synthetic ordering modes.}
\label{tab: toyexamples}
\end{table}

\paragraph{Data}
To isolate the effect of word reordering from machine translation, we ignore the word transformation from the source language to the target language. 
The {\em source sentence} is a sequence of random numbers, and the target is an ordering of the numbers (Table~\ref{tab: toyexamples}). 
To mimic the real dataset, we set the vocabulary size of numbers as 32K, and samples the sequence length from a uniform distribution from 10 to 100. 
The training set consists of 300K instances, in which the target is an ordering sampled from a given set of ordering modes from a categorical distribution.
Both the validation and test sets consist of 3K instances and all the ordering modes serves as the references for the test sets.
Suppose the training data consists of two ordering modes (e.g. direct and reverse). In testing, we will consider the output as Exact Match if the model can generate either direct or reverse orders.
This setting tries to simulate the MT scenario: although only one mode is observed in the training data, the other semantically equivalent modes should also be regarded as correct translation.

\paragraph{Model}
The NAT models used in this experiment generally share the same hyper-parameters with that in the machine translation experiment (Section~\ref{sec:mt-setup}), except that we trained the models for 30K steps, measured the validation loss at the end of each epoch, and chose the best one.

\begin{figure}[t]
    \centering
    \includegraphics[width=0.4\textwidth]{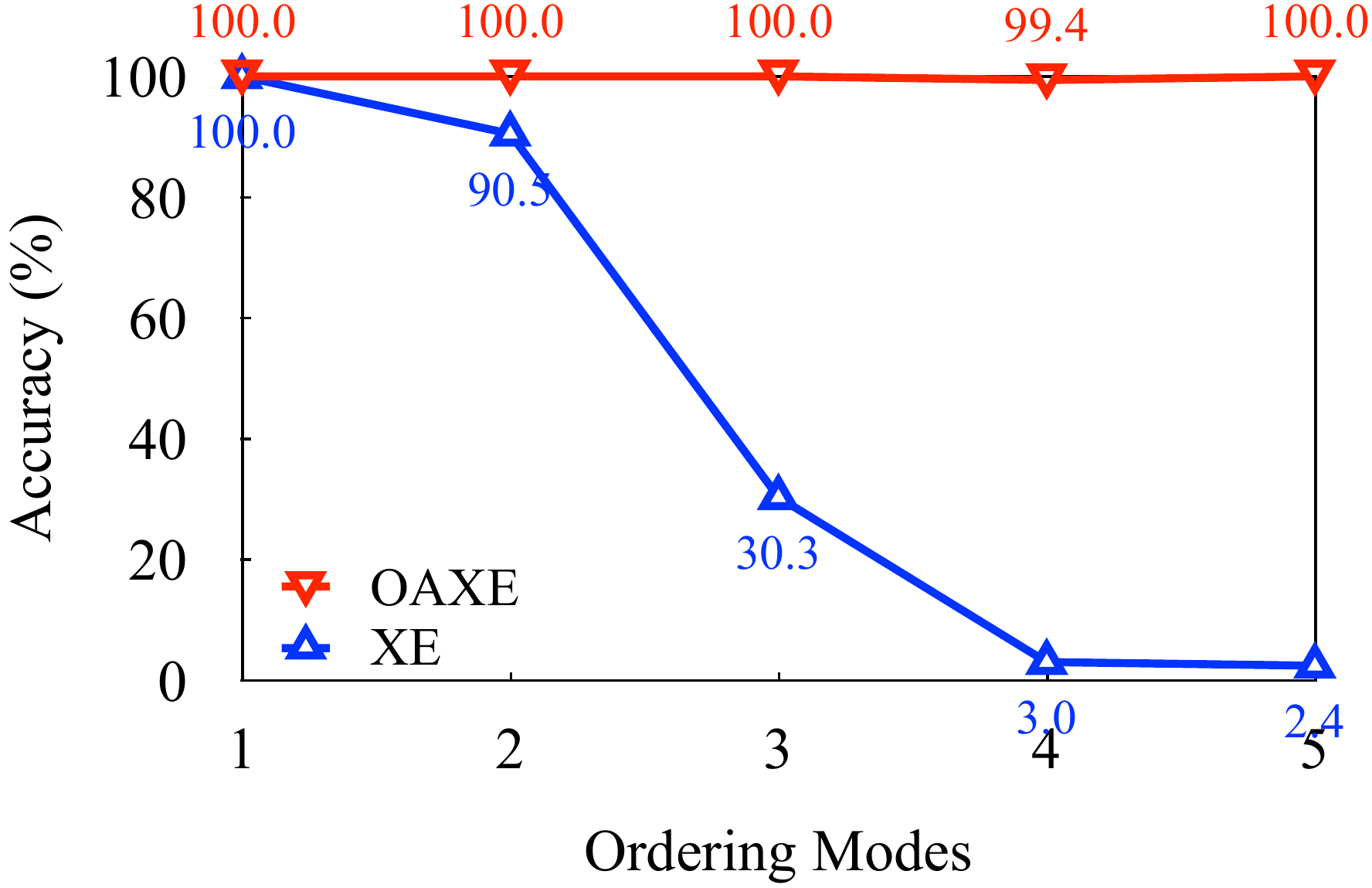}
    \caption{Results on the synthetic ordering dataset.
    ``Ordering Modes'' is the number of distinct ordering modes in the dataset.}
    \label{fig:synthetic-ordering}
\end{figure}

\paragraph{Evaluation}
Figure~\ref{fig:synthetic-ordering} shows the exact match accuracy for different number of ordering modes. ``Ordering Modes = 3'' denotes that we sample the target ordering from the set of (``Direct'', ``Reverse'', ``Flip''). 
Clearly, the performance of XE drops rapidly when the number of ordering modes increases, which reconfirms that the XE loss is weak at handling multimodality problem. Our \textsc{OaXE} performs well for multiple ordering modes, which we attribute to its superiority on finding the best ordering in the search space.

\section{Machine Translation Experiment}

\subsection{Experimental Setup}
\label{sec:mt-setup}

\paragraph{Data}
We conducted experiments on major benchmarking datasets that are widely-used in previous NAT studies~\cite{NAT,natbow,flowseq,imputer}: WMT14 English$\Leftrightarrow$German (En$\Leftrightarrow$De, 4.5M sentence pairs), WMT16 English$\Leftrightarrow$Romanian (En$\Leftrightarrow$Ro, 0.6M sentence pairs). 
We preprocessed the datasets with a joint BPE~\cite{Sennrich:BPE} with 32K merge operations for En$\Leftrightarrow$De, and 40K merge operations for En$\Leftrightarrow$Ro.
We reported the BLEU~\cite{papineni2002bleu} on both tasks.

Since our work is closely related to~\citet{maskp,axe}, we also validated our approach on the large-scale WMT17 English$\Leftrightarrow$Chinese (En$\Leftrightarrow$Zh, 20M sentence pairs) dataset. We learned a BPE model with 32K merge operations for the dataset.
For fair comparison, we reported the Sacre BLEU~\cite{post2018call} on the En-Zh task, and the BLEU on the Zh-En task.

\paragraph{Knowledge Distillation} We closely followed previous works on NAT to apply sequence-level knowledge distillation~\citep{kim2016sequence} to reduce the modes of the training data. 
We trained the NAT models on the translations generated by a left-to-right AT model. Consistent with CMLM~\cite{maskp} and AXE~\cite{axe}, we employed Transformer-\textsc{Big}~\cite{transformer} for distilling the En$\Leftrightarrow$De and En$\Leftrightarrow$Zh datasets, and Transformer-\textsc{Base} for distilling the En$\Leftrightarrow$Ro dataset.

\paragraph{NAT Models} 
We validated \textsc{OaXE} on the SOTA NAT model -- CMLM~\cite{maskp}, which uses the conditional mask LM~\citep{devlin2019bert} to generate the target sequence from the masked input.
The NAT model shares the same architecture as Transformer-\textsc{Base}: 6 layers for both the encoder and decoder, 8 attention heads, 512 model dimensions, and 2048 hidden dimensions.
We chose the CMLM models with the vanilla XE loss~\cite{maskp} and the AXE loss~\cite{axe}\footnote{Since the code of AXE was not released, we re-implemented it on top of the released CMLM code, which shares code and distilled dataset with \textsc{OaXE}. Unfortunately, our re-implementation cannot re-produce the reported BLEU scores on all benchmarks (e.g. the same on WMT14 De-En, while 1.8 BLEU lower on En-De). 
Accordingly, we re-used the reported results when available, and only used our re-implementation in Table~\ref{tab:ablation-training-strategies}.} 
as our two main baselines.
To keep consistent with main baselines, we set 5 as length candidates for all CMLM models during inference. 
We generally followed the standard hyperparameters used in~\cite{maskp}.
We trained batches of approximately 128K tokens using Adam~\citep{kingma2015adam}. 
The learning rate warmed up to $5\times10^{-4}$ in the first 10K steps, and then decayed with the inverse square-root schedule.
We trained all models for 300k steps, measured the validation BLEU at the end of each epoch, and averaged the 5 best checkpoints.

\subsection{Ablation Study}

In this section, we investigated the impact of different components for \textsc{OaXE}
on the WMT14 En$\Leftrightarrow$De validation sets.

\begin{table}[t]
    \centering
    \begin{tabular}{rcccc}
    \toprule
    \multirow{2}{*}{\bf Loss}   &  \multicolumn{2}{c}{\bf XE Annealing}    &   \multicolumn{2}{c}{\bf XE Pre-Training}\\
    \cmidrule(lr){2-3} \cmidrule(lr){4-5}
        &   \bf En-De   & \bf De-En    &   \bf En-De   & \bf De-En\\
    \midrule          
    XE          &18.6    &23.0 &18.9 &23.9 \\
    AXE  &    23.1    & 28.0  &   23.2  &  28.1\\
    \midrule
    \textsc{OaXE} &23.8 & 28.9 &\textbf{24.1 }& \textbf{29.4} \\
    \bottomrule
    \end{tabular}
    \caption{Impact of different XE initialization strategies on the WMT14 En$\Leftrightarrow$De validation set. We list the results of fine-tuning with XE (i.e., 18.9 and 23.9) to make fair comparison. We re-implemented AXE in this experiment.}
    \label{tab:ablation-training-strategies}
\end{table}

\paragraph{Different XE Initialization Strategies}
For the XE annealing experiments, we set $c$ as 16, and $\lambda$ as 0.95.
For the XE pre-training experiments, we fine-tuned all models for 10 epochs (e.g., 11K steps for En$\Leftrightarrow$De) for all language pairs, and chose the best single checkpoint with highest validation BLEU. 
The learning rate of fine-tuning warmed up to $5\times10^{-6}$ in the first 5K steps, and then decayed with the inverse square-root schedule.

Table~\ref{tab:ablation-training-strategies} shows the 
effect of different XE initialization strategies for \textsc{OaXE} training.
We compare with two XE variants which share the same model architecture, and the only difference is the training objective. 
Clearly, the proposed \textsc{OaXE} loss consistently outperforms the other XE losses in all cases, demonstrating its effectiveness for machine translation.
The XE pre-training strategy achieves the best performance, which can obtain an averaged improvement of 6.0 BLEU over the XE baseline.
In the following experiments, we use XE pre-training as the default training strategy for its efficiency and effectiveness.

However, \textsc{OaXE} fails to converge without XE initialization (i.e. 0 BLEU).
We also conducted experiments about how XE initialization with different pre-training steps affect \textsc{OaXE} in Appendix~\ref{app:pre-train}. Encouragingly, a pre-trained XE initialization at very early stage (9.0 BLEU at 10K steps) can train a reasonable \textsc{OaXE} model (18.5 BLEU).

\begin{figure}[t]
    \centering
    \includegraphics[width=0.4\textwidth]{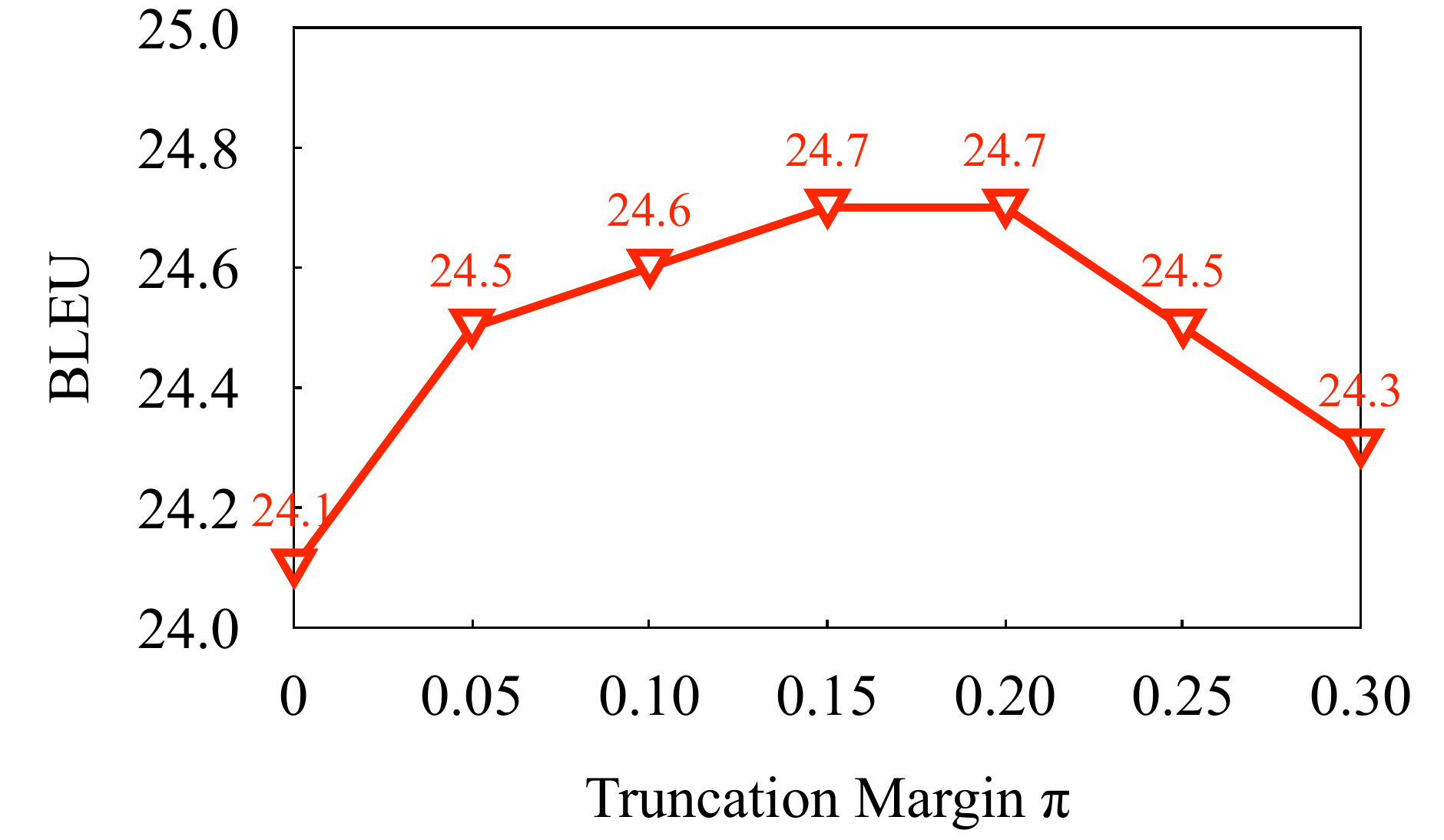}
    \caption{Impact of truncation margin on the En-De validation.}
    \label{fig:impact-of-truncation}
\end{figure}

\begin{table*}
\centering
\begin{tabular}{lcccccc}
\toprule
\multirow{2}{*}{\bf Model} & \multicolumn{2}{c}{\textbf{ WMT14 }} & \multicolumn{2}{c} { \textbf{WMT16} } & \multicolumn{2}{c} { \textbf{WMT17} }\\
 \cmidrule(lr){2-3}\cmidrule(lr){4-5}\cmidrule(lr){6-7}
 & \textbf{En-De} &\textbf{De-En} &\textbf{En-Ro} &\textbf{Ro-En} &\textbf{En-Zh}  &  \textbf{Zh-En}  \\
\midrule
\bf {Autoregressive} \\
~~~~Transformer & 27.6 & 31.4 &  34.3 &  34.0 &  34.3 & 23.7 \\
~~~~~~~~ + {Knowledge Distillation} & 27.8 & 31.3 & - & - & 34.4 &  24.0 \\
\midrule
\bf Non-Autoregressive \\
~~~~Bag-of-ngrams Loss~\cite{natbow} &20.9 &24.6 &28.3 &29.3 &  -  & -  \\
~~~~Hint-based Loss~\cite{nathint}  &21.1 &25.2 &- &-  &  -  & - \\
~~~~Flowseq~\cite{flowseq} &21.5 &26.2 &29.3 &30.4 &  -  & -  \\
~~~~Bigram CRF~\cite{natcrf} &23.4 &  27.2 &- &  -  &  -  & - \\
~~~~EM+ODD~\cite{em} & 24.5 & 27.9 & - & - & - & -\\
~~~~GLAT~\cite{glat} & 25.2 & 29.8 & 31.2 & 32.0 & - & - \\
~~~~Imputer~\cite{imputer}  &  {25.8} &28.4 &32.3 &31.7 &  -  & -  \\
\hdashline
~~~~CMLM~\cite{maskp}       &   18.3     &   22.0   &  27.6  & 28.6 & 23.6 & 13.5 \\
~~~~CMLM + AXE~\cite{axe} &   23.5     &   27.9   &  30.8   &  31.5  &  30.9   &  19.8  \\
~~~~CMLM + {\em Order-Agnostic XE} (Ours)  & \bf  ~~26.1$^{\Uparrow}$ & \bf  ~~30.2$^{\Uparrow}$ & \bf   ~~32.4$^{\Uparrow}$ & \bf   ~~33.3$^{\Uparrow}$ & \bf   ~~32.9$^{\Uparrow}$ &\bf   ~~22.1$^{\Uparrow}$  \\
\bottomrule
\end{tabular}
\caption{The performance of CMLMs trained with \textsc{OaXE}, compared to other purely non-autoregressive methods. 
We implemented our model on top of the released CMLM models~\cite{maskp}.
``$\Uparrow$'' denotes significantly better than the CMLM with $p < 0.01$.
The results of CMLM are slightly better than those reported by~\citet{maskp}, since we use de-duplication to remove repetitive tokens in the generated output.
We ue the results of AXE reported by~\citet{axe}.}
\label{tab:main}
\end{table*}

\paragraph{Impact of Truncation Margin}
\label{sec: pretrain}
Figure~\ref{fig:impact-of-truncation} shows the impact of truncation margin $\pi$, which is searched from \{$0, 0.05, 0.10, 0.15, 0.20, 0.25, 0.30$\} on the WMT14 En-De validation set.
Intuitively, the higher $\pi$ is, the more predictions are dropped. When $\pi$ increases from 0 to 0.15, we achieved 0.6 BLEU improvement by dropping likely invalid predictions. 
When $\pi$ further increases, the valid predictions can be mistakenly dropped, which leads to a performance drop.
We used $\pi=0.15$ for all language pairs in the following experiments.

\subsection{Translation Performance}

\paragraph{State of the Art}
In addition to the closely related XE variants, we also compare the performance of CMLM with \textsc{OaXE} against 7 strong baseline models: 1) bag-of-ngrams training~\cite{natbow}; 2) hint-based training~\cite{nathint}; 3) Flowseq -- a latent variable model based on generative flow~\cite{flowseq}; 4) the CRF-based semi-autoregressive model~\cite{natcrf}; 5) EM+ODD -- an EM based method to remove the multimodality in the training dataset~\citep{em}; 6) Glancing-based training~\cite{glat}; 7) Imputer -- a NAT model with latent alignments~\cite{imputer}.

We followed~\cite{nathint,em,imputer} to use de-duplication trick to remove repetitive tokens in the generated output.\footnote{Without de-duplication, the averaged BLEU score of XE is 0.07 lower, while that of \textsc{OaXE} is the same (\{25.8, 30.1, 32.2, 33.1, 32.9, 22.5\} vs \{26.1, 30.2, 32.4, 33.3, 32.9, 22.1\}). Our general conclusions still hold without de-duplication.}
Table~\ref{tab:main} shows that our approach achieves the highest BLEU scores in all benchmarks. 
Our approach achieves an averaged improvement of 2.1 BLEU over the advanced AXE on all benchmarks.
Encouragingly, \textsc{OaXE}-trained CMLMs outperforms the best fully NAT model (Imputer) by 1.0 BLEU on average, setting a new state-of-the-art for fully non-autoregressive models on the major NAT benchmarks.\footnote{When using Transformer-\textsc{Base} for distillation, the BLEU scores on WMT14 En-De and De-En are respectively 25.9 and 29.8, which still outperform previous work.}

\begin{table}
\centering
\scalebox{0.96}{
\begin{tabular}{lcc}
\toprule
\multirow{2}{*}{\bf Model} & \multicolumn{2}{c}{\textbf{ WMT14 }}\\
 \cmidrule(lr){2-3}
 & \textbf{En-De} &\textbf{De-En}\\
\midrule
CTC Loss~\cite{libovicky-helcl-2018-end}   &   17.7    &  19.8\\
Flowseq~\cite{flowseq} &  18.6    &   23.4\\
Imputer~\cite{imputer} & 15.6 &  -\\
\hdashline
CMLM~\cite{maskp}&  10.6    &   -  \\
CMLM + AXE~\citeyearpar{axe} &   20.4  &   24.9\\
\hdashline
CMLM (Our reimplemented) & 10.6 &  15.1 \\
CMLM + {\em Order-Agnostic XE} (Ours)  & \bf   ~~22.4$^{\Uparrow}$ & \bf ~~26.8$^{\Uparrow}$ \\
\bottomrule
\end{tabular}}
\caption{The performance of \textsc{OaXE}-based CMLM, compared to other non-autoregressive methods on raw data.}
\label{tab:en-de-raw}
\end{table}

\paragraph{Raw Data}
In this experiment, we evaluated the performance of fully NAT models that do not rely on any external resources (e.g., AT models for knowledge distillation). Table~\ref{tab:en-de-raw} shows the performance of NAT models that are trained on raw data without knowledge distillation.
\textsc{OaXE}-trained CMLMs still significantly outperform the other NAT models in the raw data scenarios, demonstrating the effectiveness and robustness of our approach.
Encouragingly, \textsc{OaXE} narrows the performance gap between training on the raw data and on the distilled data, moving towards independent training of NAT models without relying on external models (e.g., AT models for knowledge distillation).

\begin{table}[t]
    \centering
    \scalebox{0.96}{
    \begin{tabular}{lcrcr}
    \toprule
    \multirow{2}{*}{\bf Model}   &  \multicolumn{2}{c}{\bf Single}   &  \multicolumn{2}{c}{\bf Multiple}\\
    \cmidrule(lr){2-3} \cmidrule(lr){4-5}
        &  \bf BLEU & \bf $\Delta$    &  \bf BLEU & \bf $\Delta$\\
    \midrule
    \bf Raw Data  \\
    
    ~~~~CMLM                  &11.0 &   -    &  28.1 & - \\
    ~~~~CMLM + \textsc{OaXE}     &  22.7 &  +11.7 &  57.5 & \bf +29.4 \\
    \midrule     
    \bf Distillation  \\
    
    ~~~~CMLM                  &18.6  &  -     & 50.7 &  -  \\
    ~~~~CMLM + \textsc{OaXE}     &26.2  &  +7.6  & 68.0 &  \bf +17.3 \\
    \bottomrule
    \end{tabular}}
    \caption{BLEU scores on WMT14 En-De test sets with different number of references (e.g., ``Single'' and ``Multiple''). ``$\Delta$'' denotes the improvement of \textsc{OaXE} over the vanilla CMLM model.}
    \label{tab:reference}
\end{table}

\paragraph{Multiple References}
As aforementioned, the multimodality problem that NAT models suffer is mainly due to the uncertainty nature of translation. In this experiment, we measure how well NAT models can handle the translation uncertainty using multiple semantically equivalent references.
Specifically, we use the dataset released by~\citet{ott2018analyzing} for evaluating translation uncertainty, which consists of ten human translations for 500 sentences taken from the WMT14 En-De test set.
Table~\ref{tab:reference} lists the results. Two observations can be made.
Firstly, \textsc{OaXE} achieves more significant improvement over the vanilla XE when measured by multiple references. One possible reason is that \textsc{OaXE} produces more diverse translations with different word orders, which can be better measured by multiple references.

Secondly, when measured by multiple references, the improvement of \textsc{OaXE} on the raw data is even more significant than that on the distilled data.
This is intuitive, since the raw data is more complex in terms of ``modes'' (alternative translations for an input), and knowledge distillation reduces the modes to help NAT to model the variations in the output data~\cite{zhou2019understanding,Ding:2021:ACL,Ding:2021:ICLR}.
Accordingly, knowledge distillation is essential for the XE-based NAT, which is weak at modeling multimodality. In contrast, \textsc{OaXE} reduces the modes by relaxing the order constraint, thus releases the reliance on the knowledge distillation.

\begin{figure}[t]
    \centering
    \subfloat[Raw Data]{
    \includegraphics[height=0.33\textwidth]{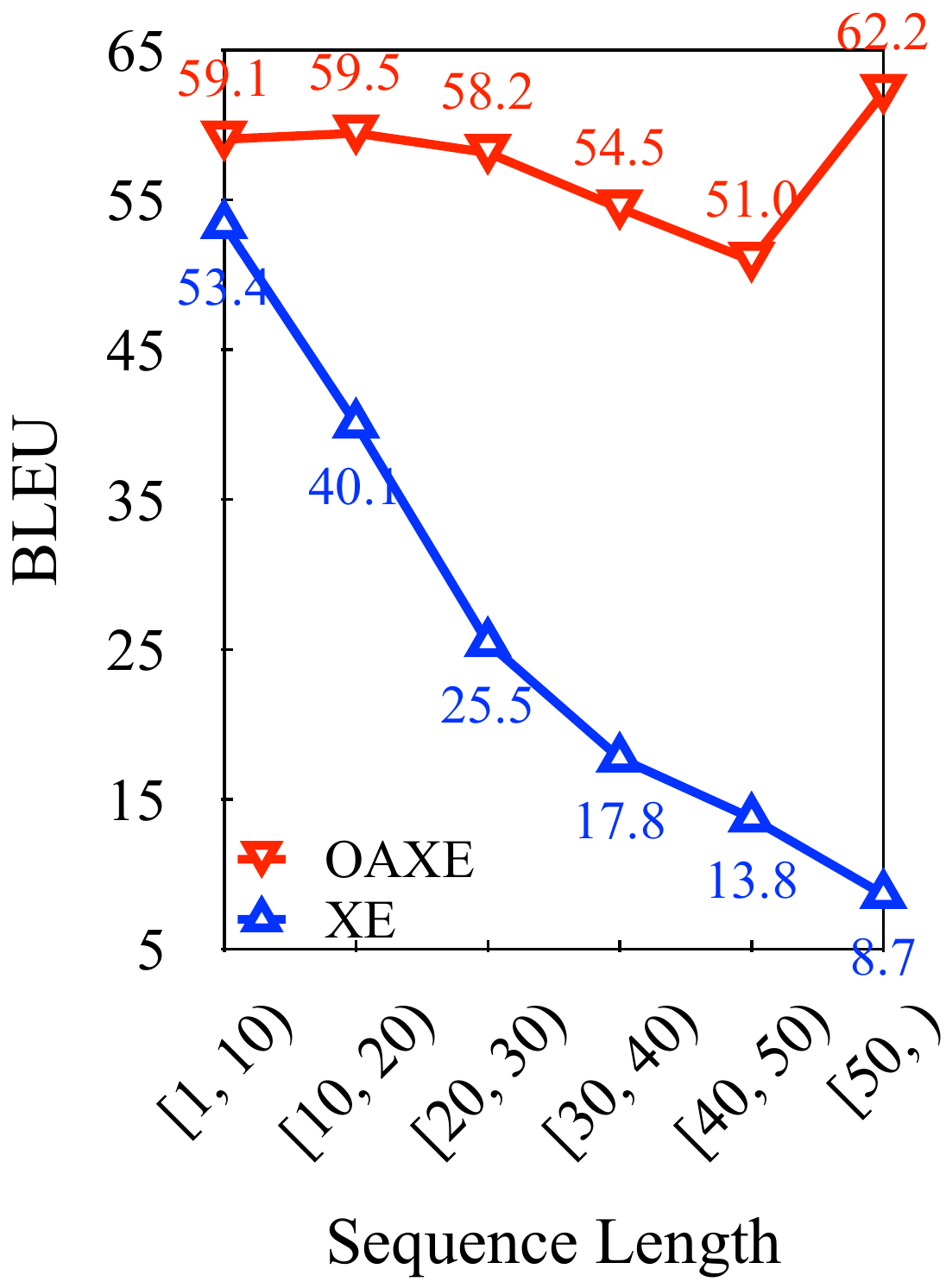}} \hfill
    \subfloat[Knowledge Distillation]{
    \includegraphics[height=0.33\textwidth]{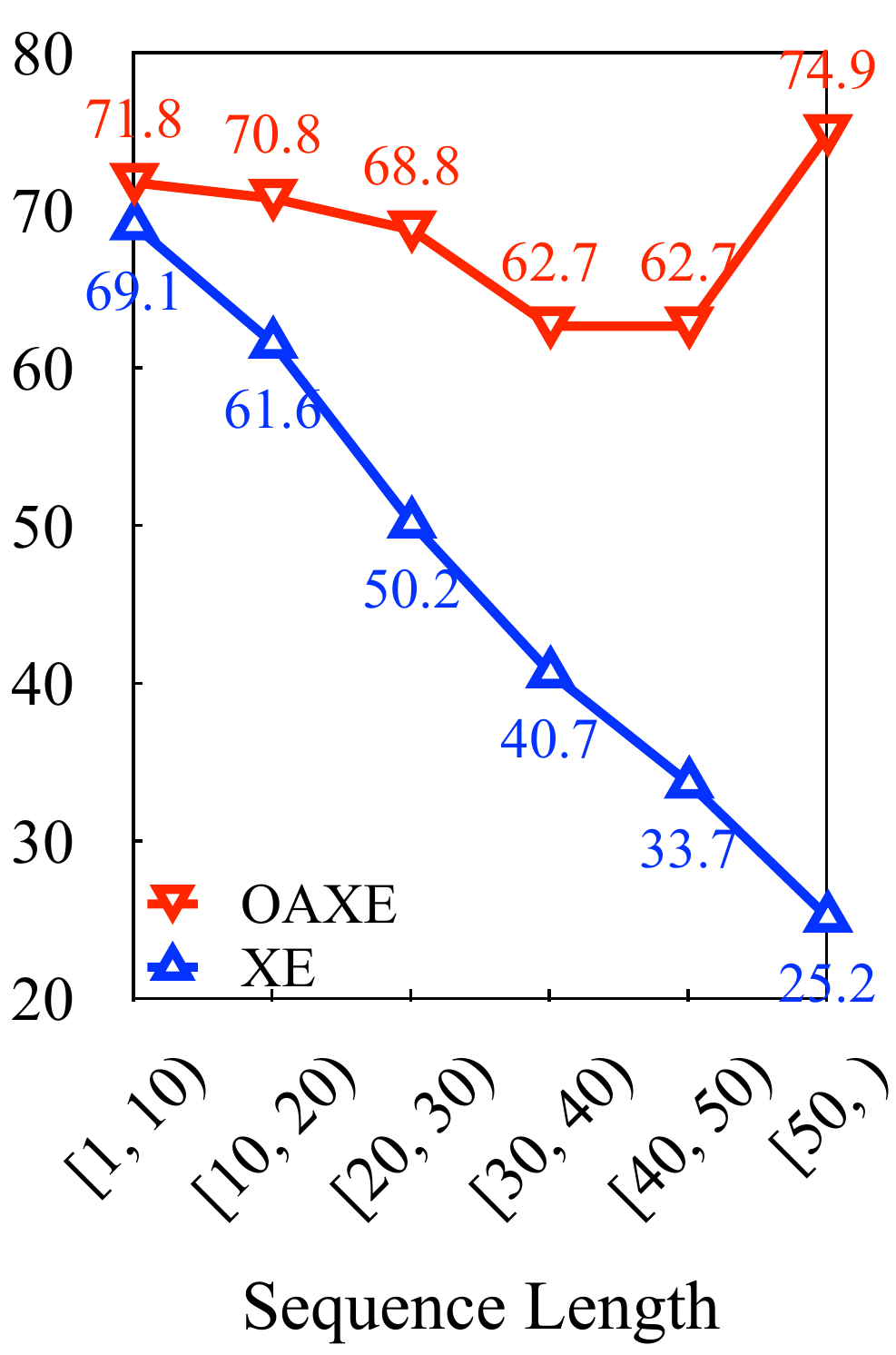}}
    \caption{Performance of the generated translations with respect to the lengths of the reference sentences. Results are reported on the WMT14 En-De test sets with multiple references (e.g., Table~\ref{tab:reference}), which can better measure the quality of the generated outputs.}
    \label{fig:sentence-length-multi}
\end{figure}

\paragraph{Sequence Lengths} 
We also investigate the model performance for different sequence lengths. We split the test sets into different buckets based on the reference sentence length, indicating whether a system does better or worse at shorter or longer sentences.
Figure~\ref{fig:sentence-length-multi} shows that results on the sampled WMT14 En-De test set with multiple references (i.e., Table~\ref{tab:reference}), which can better measure the translation quality.
As seen, the performance of XE drops rapidly when the sequence length increases. 
Our \textsc{OaXE} significantly improves performance, especially on longer sentences by further relaxing the strict order constraint, which is especially important for handling long sentences.
We also compare \textsc{OaXE} with the vanilla XE and AXE on the full test set of WMT14 En$\Leftrightarrow$De (Appendix~\ref{sec:app-length}), in which \textsc{OaXE} consistently outperforms both XE and AXE.

\begin{table}
\centering
\begin{tabular}{lc}
\toprule
\bf Model   &   \bf Ja-En   \\
\midrule
Transformer-Base    &   28.5\\
\midrule
CMLM (Our reimplemented)       & 20.9\\
CMLM + {\em Order-Agnostic XE} (Ours) &  28.4\\
\bottomrule
\end{tabular}
\caption{Performance on WAT17 Japanese-English data.}
\label{tab:ja-en}
\end{table}

\paragraph{Syntactically Divergent Language Pair}
To investigate whether our approach would work for more syntactically divergent language pairs, we conducted experiments on the Japanese-English pair, which differ greatly in the word orders (subject-verb-object language vs. subject-object-verb language).
Specifically, we used the first two sections of WAT17 Japanese-English dataset, which consists of 1.9M sentence pairs. We employed Transformer-\textsc{Base} model for distilling the training data.
Table~\ref{tab:ja-en} shows the results on the WAT17 Japanese-English data. \textsc{OaXE} significantly improves over the fully CMLM model from 20.9 to 28.4 with a Transformer-Base teacher (28.5).

\subsection{Analysis}
\label{sec:analysis-output}

In this section, we present a qualitative analysis to provide some insights where \textsc{OaXE} improves over cross entropy by alleviating the multimodality problem.
To better understand the proposed \textsc{OaXE}, we report the results for the two newly introduced components -- \textsc{OaXE} loss and loss truncation.
To make fair comparison with previous work~\cite{maskp,axe}, we analyze the generated outputs of the full WMT14 En$\Leftrightarrow$De test sets (e.g., Tables~\ref{tab:main} and~\ref{tab:en-de-raw}).

\begin{table}[t]
    \centering
    \begin{tabular}{lrr}
    \toprule
    \multirow{2}{*}{\bf Model}   &  \multicolumn{2}{c}{\bf WMT14}\\
    \cmidrule(lr){2-3}
        &   \bf En-De   & \bf De-En\\
    \midrule
    \bf Gold Test Set   &   0.04\%  &   0.02\%\\
    \midrule
    \bf Raw Data  \\
    
    ~~~~CMLM                &22.80\%   &22.70\%\\
    ~~~~CMLM + \textsc{OaXE} Loss & 1.70\% & \bf 1.39\%\\
    ~~~~~~~ + Loss Truncation     &  \bf 1.57\% & 1.67\%\\
    \midrule
    \bf Knowledge Distillation  \\
    
    ~~~~CMLM                           &   16.72\% &   12.31\%\\
    ~~~~CMLM + \textsc{OaXE} Loss      &  0.90\% &  0.88\% \\
    ~~~~~~~ + Loss Truncation          & \bf 0.81\% & \bf 0.73\% \\
    \hdashline
    ~~~~CMLM + AXE~\citeyearpar{axe}   &  1.41\%   &   1.03\%  \\
    \bottomrule
    \end{tabular}
    \caption{Repeated token percentage on the test sets. 
    The results of AXE are reported by~\citet{axe}.}
    \label{tab:repetition}
\end{table}

\paragraph{\textsc{OaXE} Reduces Token Repetitions}
Repeated token percentage is a commonly-used metric of measuring multimodality in a NAT model~\cite{axe,imputer}.
A NAT model may consider many possible translations at the same time due to the independent predictions of target words.
Accordingly, the NAT output typically contains many repetitive tokens. 

Table~\ref{tab:repetition} shows the repetition rate for models trained on both distilled and raw datasets. We also list the results of AXE reported by~\cite{axe} for reference.
\textsc{OaXE} significantly reduces the repetition percentage over the vanilla XE by a multiplicative factor of 13 on both raw and distilled data. In addition, the proposed \textsc{OaXE} outperforms AXE in token repetition, which is consistent with the translation results in Tables~\ref{tab:main} and~\ref{tab:en-de-raw}.

\begin{table}[t]
    \centering
    \begin{tabular}{lrr}
    \toprule
    \multirow{2}{*}{\bf Model}   &  \multicolumn{2}{c}{\bf WMT14}\\
    \cmidrule(lr){2-3}
        &   \bf En-De   & \bf De-En\\
    \midrule
        \bf Raw Data  \\
    ~~~~CMLM                &   1.30 &   0.64\\
    ~~~~CMLM + \textsc{OaXE} Loss & 1.00  & 0.55 \\ 
    ~~~~~~~ + Loss Truncation & \bf 0.33 & \bf 0.29\\    
    \midrule
    \bf Knowledge Distillation   \\
    ~~~~CMLM                &   0.55 &   0.47\\
    ~~~~CMLM + \textsc{OaXE} Loss &  0.30  & 0.27 \\
    ~~~~~~~ + Loss Truncation     & \bf 0.19 & \bf 0.19 \\
    \bottomrule
    \end{tabular}
    \caption{Normalized Corpus-level multimodality (NCM) scores of the generated outputs on the test sets.}
    \label{tab:NCM}
\end{table}

\paragraph{\textsc{OaXE} Increases Prediction Confidence}
Conditional entropy is another metric of measuring corpus-level multimodality for NAT model~\cite{em}.
Since a NAT model considers many possible translations at the same time, it tends to spread too
much probability mass over the space of sequences and thus generates translations which have lower generation probabilities.
Formally, given a bilingual corpus ($\mathcal{X}$, $\mathcal{Y}$), the Normalized Corpus-level multimodality (NCM)~\cite{em} is calculated as
\begin{equation*}
\operatorname{NCM}_{\mathcal{X}}(\mathcal{Y})=\frac{\mathbb{E}_{(X, Y) \sim(\mathcal{X}, \mathcal{Y})}\left[-\log P\left(Y \mid X \right)\right]}{\mathbb{E}_{Y \sim \mathcal{Y}}[|Y|]},
\end{equation*}
where $|Y|$ is the length of the output sequence $Y$.

We use NCM to compare the outputs generated by different NAT models on the same test set, as listed in Table~\ref{tab:NCM}. The proposed \textsc{OaXE} consistently outperforms \textsc{XE} across language pairs and datasets, which proves that \textsc{OaXE} can increase output confidence by better estimating the probability of generated output in potentially different orders.

\begin{table*}[t]
    \caption{Example for De$\Rightarrow$En translation. CMLM often generates {\color{zred} repetitive tokens}, while \textsc{OaXE} suffers less from these mistakes.}
    \label{tab:case-study}
    \centering
    \begin{tabular}{l m{14.5cm}}
    \toprule
    \bf Source & Passagiere schimpfen häufig über Gepäckzuschläge und andere Gebühren , doch Fluggesellschaften greifen gern darauf zurück .\\
    \bf Reference & Passengers often grumble about baggage charges and other fees , but airlines love them.\\
    \hdashline
    \bf CMLM & Passengers often abuse at baggage {\color{zred} baggage} charges {\color{zred} charges} and charges {\color{zred} charges} , airlines {\color{zred} airlines} like to use them .\\
    \bf ~~~+ \textsc{OaXE}  & Passengers often complain about baggage charges and other charges , but airlines like to resort to them .\\
    \bottomrule
    \end{tabular}
\end{table*}

\begin{table}[t]
    \centering
    \begin{tabular}{lrr}
    \toprule
    \multirow{2}{*}{\bf Model}   &  \multicolumn{2}{c}{\bf WMT14}\\
    \cmidrule(lr){2-3}
        &   \bf En-De   & \bf De-En\\
    \midrule
    \bf Gold Test Set   &   37.0 & 43.8   \\
    \midrule
    \bf Raw Data  \\
    ~~~~CMLM         & 511.6 & 351.8 \\
    ~~~~CMLM + \textsc{OaXE} Loss  &  146.9     & 120.1  \\
    ~~~~~~~ + Loss Truncation      & \bf 123.1 & \bf 105.4 \\
    \midrule
    \bf Knowledge Distillation   \\
    
    ~~~~CMLM                                   &  205.6    & 177.6 \\
    ~~~~CMLM + \textsc{OaXE} Loss   &   82.6    & 83.7    \\
    ~~~~~~~ + Loss Truncation       & \bf 69.4  & \bf 77.2 \\
    \bottomrule
    \end{tabular}
    \caption{PPLs of the generated outputs on the test sets. Lower PPL score denotes better fluency.}
    \label{tab:fluency}
\end{table}

\paragraph{\textsc{OaXE} Improves Generation Fluency}
Some researchers may doubt that removing the order constraint in the \textsc{OaXE} loss will make NAT models generate ungrammatical output (e.g., correct predictions in wrong positions). 
In response to this problem, we measure the fluency with language models released by Fairseq,
which are trained on the News Crawl corpus for the target language.
To better evaluate the fluency of the generated output, we use a practical trick {\em de-duplication}~\cite{imputer} to remove the repetitive tokens.

Results in Table~\ref{tab:fluency} can {dispel} such doubt by showing that \textsc{OaXE} consistently improves fluency in all cases. We attribute the improvement of fluency to that XE initialization can avoid a large portion of invalid orderings in the search space, and the loss truncation can drop invalid predictions to further improve the fluency of the searched ordering.
Encouragingly, \textsc{OaXE} trained on the raw data outperforms the standard XE trained on the distilled data, showing that \textsc{OaXE} has the ability to directly learn from the raw data of complex order modes.

\paragraph{Case Study} Table~\ref{tab:case-study} shows a translation example, where the vanilla CMLM output contains many repetitive tokens due to suffering from severe multimodality problem. In addition, it also misses some important function words like ``other'' and ``but'', which harm the output fluency. The proposed \textsc{OaXE} can successfully correct these mistakes.

\section{Related Work}

Compared with AT models, NAT breaks the conditional dependencies among target tokens, which greatly degrades the model performance due to the multimodality problem.
To bridge the performance gap between NAT and AT models, several approaches try to build dependencies among the target tokens. For example,~\citet{natbow} use bag of n-grams loss to capture the target-side sequential dependency, and~\citet{flowseq} model the joint distribution of all predictions with generative flow.~\citet{Shu2020LaNMT} introduce a sequence of continuous latent variables to capture the uncertainty in the target sentence.
In this work, we improve the cross entropy loss to ameliorate the effect of multimodality by better handling word reordering.
Experimental results show that our approach consistently outperforms previous studies on alleviating the multimodality problem, and set new state of the art on the major WMT benchmarks.

Knowledge distillation is a crucial early step in the training of almost all existing NAT models.
~\citet{NAT} claim that NAT suffers from the multimodality problem (i.e., multiple lexical translations for a source word), and knowledge distillation can simplify the dataset, which is empirically validated by~\citet{zhou2019understanding}.
~\citet{ren2020astudy} reveal that knowledge distillation reduces the token dependency in target sequence and thus improves the accuracy of NAT models.
\textsc{OaXE} narrows the performance gap training on raw data and on the distilled data for its superiority on handling multimodality,
which opens up the possibility that NAT models can be trained directly on the raw data without knowledge distillation.

\section{Conclusion}

In this work, we propose a new training objective \textsc{OaXE} to handle word reordering in translation, which is one common source to the multimodality problem.
We validated our approach on six major WMT benchmarks, and showed that \textsc{OaXE} significantly improves translation performance over the standard \textsc{XE}-based NAT models, which closely match the performance of AT models.
\textsc{OaXE}-based NAT also narrows the performance gap between training on raw data and on the distilled data, indicating the potential to train purely non-autoregressive models without knowledge distillation.
Future directions include validating \textsc{OaXE} on other tasks (e.g., dialogue and summarization) and other NAT models.

\section*{Acknowledgements}

We thank all anonymous reviewers for their insightful comments. 
This project is jointly supported by Rhino-Bird Elite Training Program of Tencent AI Lab and the National Research Foundation, Singapore under its International Research Centres in Singapore Funding Initiative. Any opinions and conclusions or recommendations expressed in this material are those of the authors and do not reflect the views of National Research Foundation, Singapore.

\bibliography{ref}
\bibliographystyle{icml2021}

\clearpage

\appendix

\section{Appendix}

\subsection{Pytorch-Style Pseudo-Code of \textsc{OaXE}}
\label{sec:app-pseudocode}

\begin{algorithm}[h]
   \caption{Order Agnostic Cross Entropy}
   \label{alg:pseudocode}
\begin{algorithmic}
   \STATE {\bfseries Input:} Ground truth $Y$, NAT output log predictions $P$
   \STATE $bs$, $length$ = $Y$.size()
   \STATE $Y$ = $Y$.repeat(1, $length$).view($bs$, $length$, $length$)
   \STATE $costMatrix$ = -$P$.gather(index=$Y$,~dim=2)
  \FOR{$i=0$ {\bfseries to} $bs$}
  \STATE $bestMatch$[i] = HungarianMatch($costMatrix$[i])
  \ENDFOR
   \STATE {\bfseries Return:}$costMatrix$.gather(index=$bestMatch$,~dim=2)
\end{algorithmic}
\end{algorithm}

\subsection{Details for Synthetic Ordering Experiment}
\paragraph{Distribution} The categorical distributions for synthetic ordering experiment are randomly sampled from the Dirichlet Distribution, we list these distributions at following:
\begin{itemize}
    \item 2 Modes: [0.53, 0.47]
    \item 3 Modes: [0.23, 0.44, 0.33]
    \item 4 Modes: [0.17, 0.28, 0.14, 0.41]
    \item 5 Modes: [0.14, 0.25, 0.13, 0.39, 0.09]
\end{itemize}
\paragraph{Inference} Due to this experiment only cares about the multimodality of word orders, to erase the error caused by length predictor, for all CMLM models we decoding with the reference target length.

\subsection{Impact of Pre-Trained XE-Based NAT Models}
\label{app:pre-train}

\begin{figure}[h]
    \centering
    \includegraphics[width=0.4\textwidth]{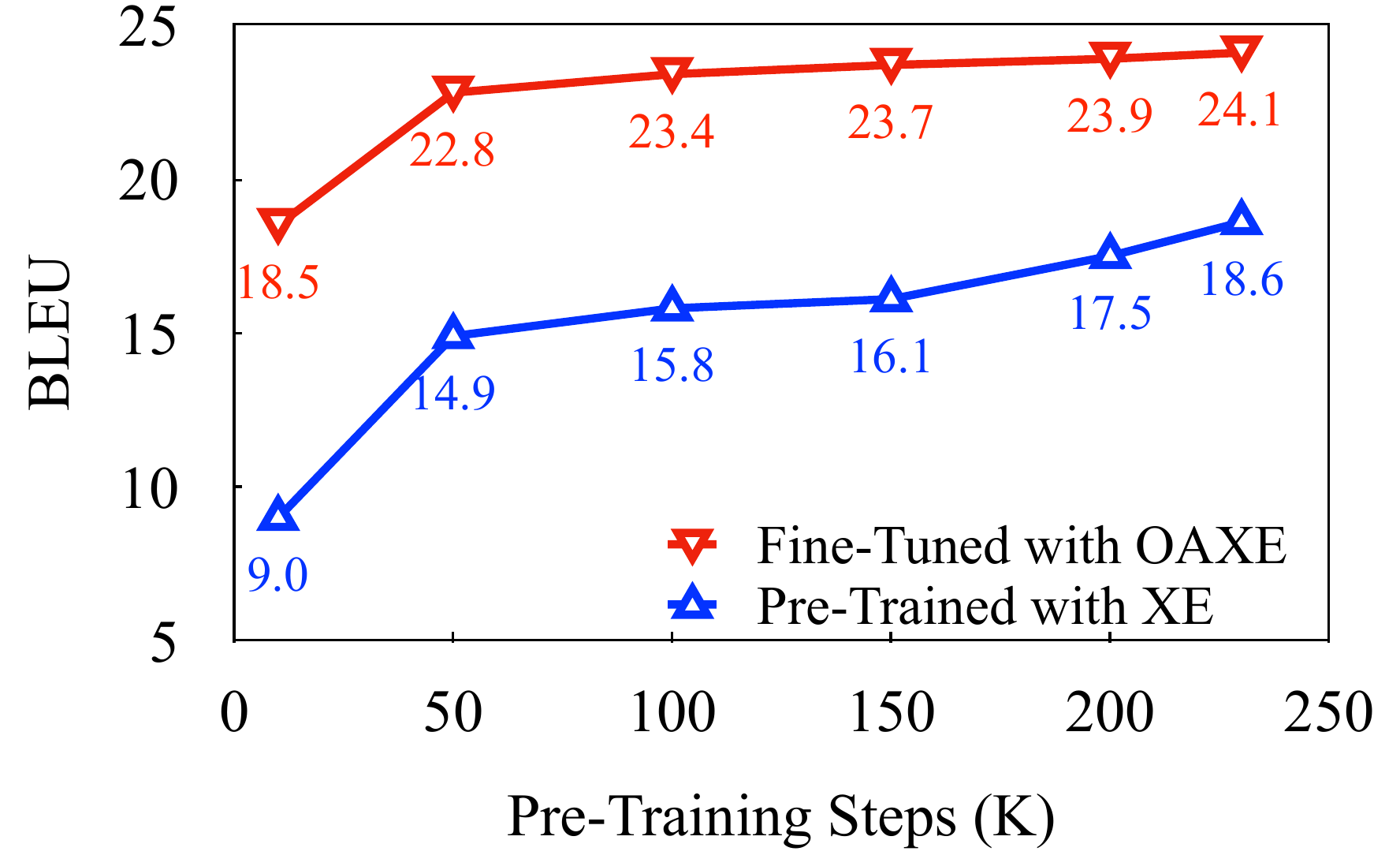} 
    \caption{Impact of pre-trained NAT models with different training steps on the WMT14 En-De validation set.}
    \label{fig:impact-of-pre-trained}
\end{figure}

Figure~\ref{fig:impact-of-pre-trained} shows the impact of different XE pre-training steps on the WMT14 En-De validation set.
For each pre-trained model, we fine-tune with \textsc{OaXE} for 10 more epochs. \textsc{OaXE} consistently and significantly improves performance at all steps, demonstrating its robustness. Encouragingly, given a pre-trained NAT at early stage (e.g., 15.8 at 100K steps), fine-tuning with \textsc{OaXE} achieves 23.4 BLEU with 11K more training steps, which outperforms both the aligned and vanilla XEs with fewer training steps (111K vs 230K).

\subsection{Different Masking Objectives}

\begin{table}[h]
    \centering
    \scalebox{0.95}{
    \begin{tabular}{ccccc}
    \toprule
    \multicolumn{2}{c}{\bf Training Objective}   &  \multicolumn{2}{c}{\bf WMT14}\\
    \cmidrule(lr){1-2} \cmidrule(lr){3-4}
    \bf Input Tokens    &   \bf Loss Functions  &   \bf En-De   &   \bf De-En\\
    \midrule
    \em Unobserved  &   \em All Tokens  &   24.1    &   29.4\\
    \em Partially-Observed  &   \em All Tokens  &  24.0 & 29.1 \\
    \em Partially-Observed  &   \em Only Masks  &  24.0 & 29.0 \\
    \bottomrule
    \end{tabular}}
    \caption{Ablation  study of different masking strategies for \textsc{OaXE} on WMT14 En$\Leftrightarrow$De validation set.}
    \label{tab:masking}
\end{table}

Due to the flexibility of CMLM, there are multiple different training mask strategies. However, we found our method is not sensitive to the mask strategies as shown in Table~\ref{tab:masking}. 
The strategy \textit{Mask All and Predict All Tokens} achieves best performance and we attribute it to the consistency of training and inference for purely non-autoregressive translation.  Due to its simpleness and effectiveness, we choose it as the default setting.

\subsection{Results of Sequence Lengths on Full Test Sets}
\label{sec:app-length}

\begin{figure}[h]
    \centering
    \includegraphics[width=0.38\textwidth]{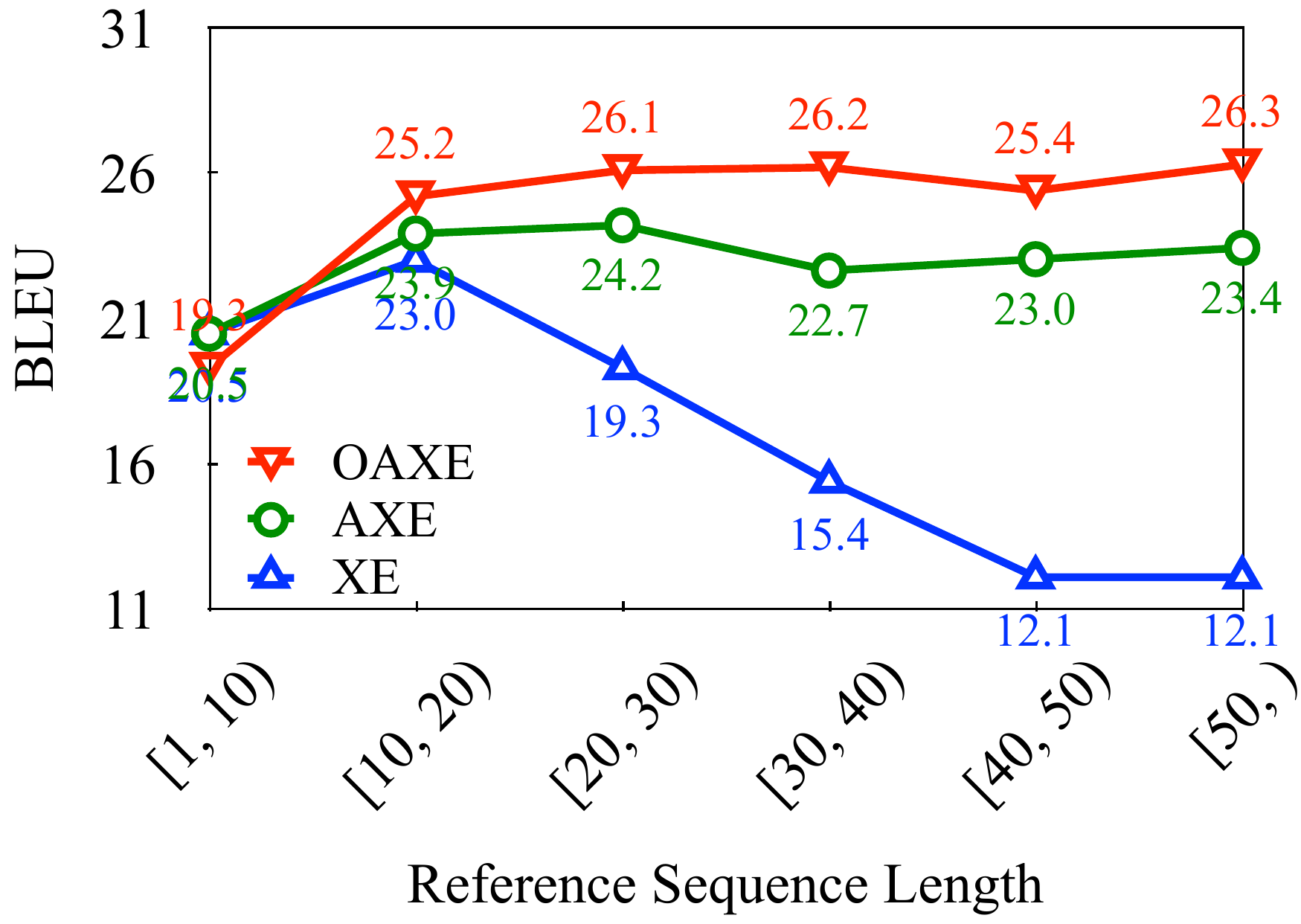} \\ \vspace{10pt}
    \includegraphics[width=0.38\textwidth]{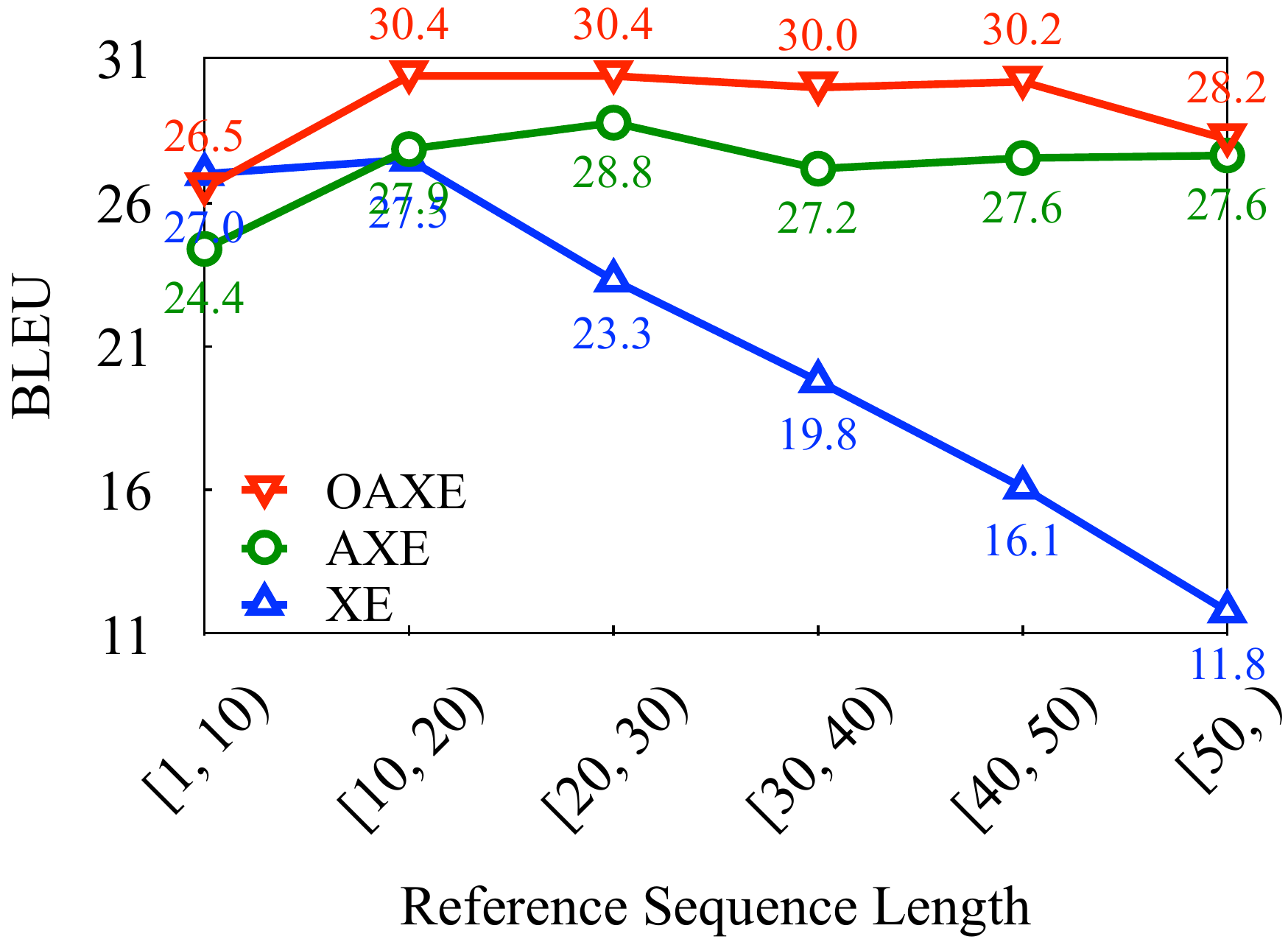}
    \caption{Performance of different sequence lengths on the distilled data for WMT14 En-De (upper panel) and De-En (bottom panel).}
    \label{fig:sentence-length}
\end{figure}

\clearpage

\end{document}